\documentclass[3p,preprint,times,nolinenumbers]{elsarticle}

\usepackage{hyperref}
\usepackage{float}
\usepackage{amsmath}
\usepackage{amsfonts}
\usepackage{amssymb}
\usepackage{graphicx}
\usepackage{url}

\begin{document}

\begin{frontmatter}



\title{Bridging the Gap: Doubles Badminton Analysis with Singles-Trained Models}


\author[ssu1]{Seungheon Baek}
\author[ssu2]{Jinhyuk Yun\corref{cor1}}
\ead{jinhyuk.yun@ssu.ac.kr}

\address[ssu1]{Department of Intelligent Semiconductors, Soongsil University, Seoul 06978, Republic of Korea}
\address[ssu2]{School of AI Convergence, Soongsil University, Seoul 06978, Republic of Korea}

\cortext[cor1]{Corresponding author}

\begin{abstract}
Badminton is known as one of the fastest racket sports in the world. Despite doubles matches being more prevalent in international tournaments than singles, previous research has mainly focused on singles due to the challenges in data availability and multi-person tracking. To address this gap, we designed an approach that transfers singles-trained models to doubles analysis. We extracted keypoints from the ShuttleSet single matches dataset using ViT-Pose and embedded them through a contrastive learning framework based on ST-GCN. To improve tracking stability, we incorporated a custom multi-object tracking algorithm that resolves ID switching issues from fast and overlapping player movements. A Transformer-based classifier then determines shot occurrences based on the learned embeddings. Our findings demonstrate the feasibility of extending pose-based shot recognition to doubles badminton, broadening analytics capabilities. This work establishes a foundation for doubles-specific datasets to enhance understanding of this predominant yet understudied format of the fast racket sport.
\end{abstract}


\end{frontmatter}


\section{Introduction}\label{sec:intro}

Understanding the dynamics of human motion and coordination in competitive sports has long been a central focus in sports science and biomechanics. Traditionally, this field has relied primarily on limited quantitative analysis based on manual observation, where researchers carefully examine individual movements and categorize them into tactical frameworks. However, recent advances in computer vision and machine learning have fundamentally transformed how scholars can address such complex behavioral inquiries in sports analytics~\cite{thomas2017computer,fujii2025computer}. The unprecedented availability of high-quality video data, combined with novel analytical frameworks including deep learning, pose estimation, and tracking algorithms~\cite{yolo,botsort,xu2022vitpose,stgcn,transformer}, has opened new avenues for systematic athletic performance analysis.

\begin{figure}[]
    \centering
    \includegraphics[width=0.7\textwidth]{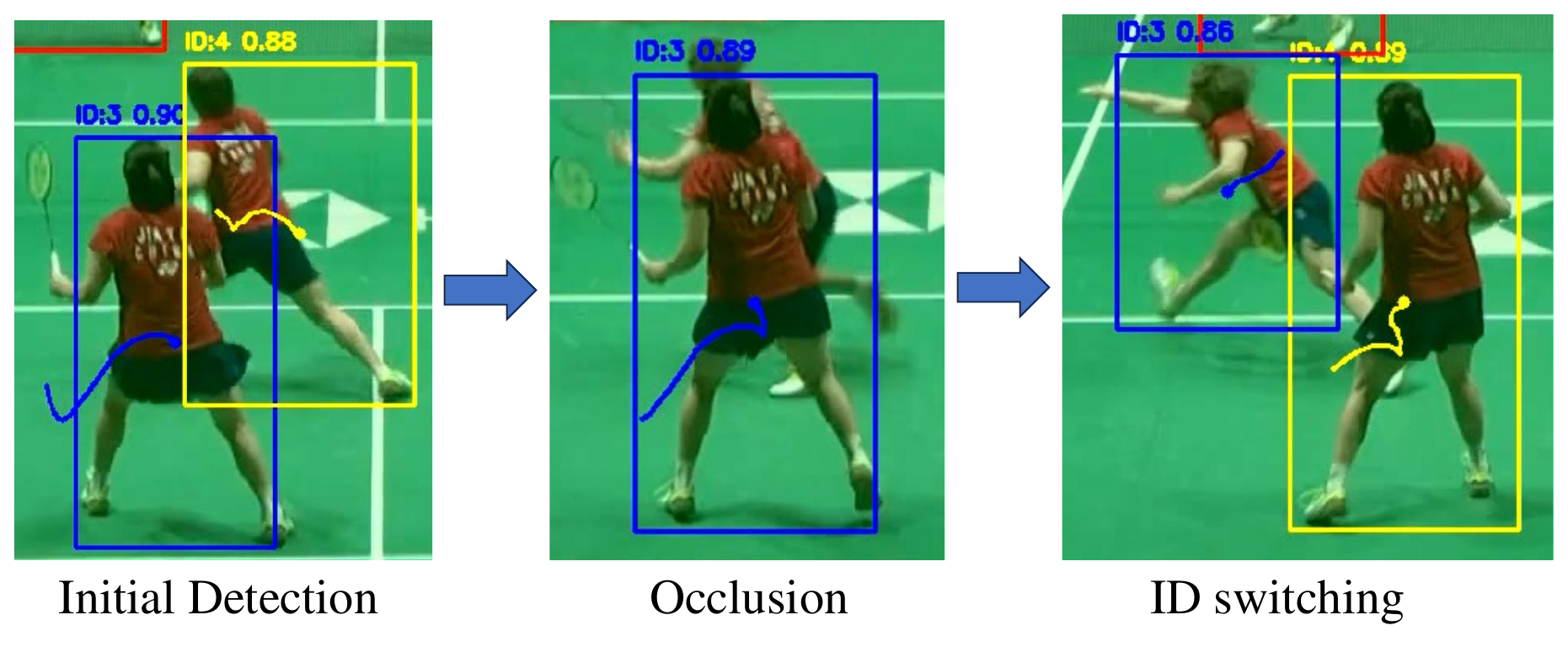}
    \caption{
        Example of object detection and identity switching in doubles badminton.
        \textbf{(Left)} Both players are correctly detected and assigned consistent IDs.
        \textbf{(Middle)} One player becomes partially occluded due to overlap, increasing the risk of tracking errors.
        \textbf{(Right)} After the occluded player reappears, an ID switch occurs, assigning a new identity. 
        Such failures in consistent tracking hinder accurate shot recognition, as the model may associate an action with the incorrect player.}
    \label{fig:tracking_issue}
\end{figure}

In this context, badminton exemplifies a particularly challenging domain for such studies due to its notable characteristics: high-speed shuttlecock trajectories, short but intense rallies, and demands for rapid tactical decisions. Among its diverse game styles, the doubles game presents significantly greater analytical complexity than singles, where four players, two for each side, perform within confined spatial boundaries. This format involves cooperative strategies, dynamic positional rotations, and synchronized movement patterns that are fundamentally distinct from singles play, significantly increasing analytical complexity. In addition, doubles play represents the predominant format of badminton competition; for example, it constitutes three of the five disciplines in the Badminton World Federation (BWF) tournament structure: men's doubles, women's doubles, and mixed doubles. However, most of the computational studies have concentrated on singles matches, leaving doubles play relatively underexplored compared to singles~\cite{cuiping2021badminton,luo2022vision,rahmad2020vision}. This gap consequently results in a lack of appropriate datasets, limiting the development of computational methods for doubles. 

The inherent difficulties in doubles badminton are multifaceted. Primarily, a doubles game requires simultaneous multi-person tracking, yet frequent player occlusions occur due to overlapping movements and rapid positional exchanges during rallies. Additionally, because players on the same side typically wear identical uniforms, ambiguous visual cues also confuse detection and identification processes. In this challenging context, identity switching (ID switching), where tracking algorithms incorrectly reassign player identities during temporary occlusions or rapid movements, emerges as a particularly critical issue. This technical limitation imposes significant constraints on downstream analytical tasks, as pose-based shot recognition systems depend on temporally consistent identity assignment to associate specific movements with the correct player.

Figure~\ref{fig:tracking_issue} illustrates this challenge through a representative example. Initially, both players are accurately detected and consistently tracked (left). However, as players converge spatially, visual occlusion occurs, compromising the tracking algorithm's ability to maintain identity consistency (middle). Upon reappearance, the previously occluded player may be assigned a new identity, breaking the temporal continuity essential for reliable pose sequence analysis (right). Such failures not only disrupt individual player tracking but also amplify errors throughout the analytical pipeline.

To address these fundamental challenges, we propose a novel computational framework that adapts a pose-based shot recognition model, originally trained on singles play, to accommodate the complex multi-player dynamics of doubles badminton. Our approach provides robust shot event detection while maintaining tracking quality sufficient to preserve identity consistency across temporal sequences, ensuring that pose trajectories remain correctly attributable to individual players.

Our methodological pipeline encompasses three interconnected components: (1) enhanced player detection and multi-object tracking with identity preservation mechanisms, (2) spatio-temporal pose sequence embedding for movement characterization, and (3) shot event classification using deep learning architectures. 

In the detection and tracking stage, we employ YOLOv11x~\cite{yolo} for precise player localization combined with BoT-SORT~\cite{botsort} for temporal association. Recognizing the limitations of standard tracking approaches in high-density scenarios, we introduce a lightweight post-processing strategy that predicts player positions during temporary disappearances and reassigns original identities through simple extrapolation based on object velocity. The core contribution lies in our shot recognition architecture, which integrates ViT-Pose~\cite{xu2022vitpose} for accurate keypoint extraction with spatio-temporal analysis frameworks. We aggregate frame-wise pose vectors into fixed-length sequences capturing movement dynamics, then apply ST-GCN~\cite{stgcn} with contrastive learning objectives to distinguish shot-related movements from routine positional adjustments. Finally, a Transformer encoder~\cite{transformer} classifies pose sequences, leveraging self-attention mechanisms to capture long-range temporal dependencies crucial for analyzing rapid, complex movements in doubles play.

By combining minimal yet effective ID-preserving tracking with advanced spatiotemporal pose learning, our framework extends shot recognition capabilities to the previously underexplored doubles setting. Experimental results demonstrate that our method achieves robust performance despite visual clutter and occlusion. Our work thus bridges the gap by providing the first systematic cross-format transfer learning approach in racket sports, demonstrating that pose-based models achieve 2$\times$ better transferability than vision-only methods. More importantly, this work lays the foundation for doubles-specific datasets and fine-grained rally analysis—an essential step toward intelligent sports analytics in multi-player scenarios. 

\section{Related Work}\label{sec:rel_work}
Badminton poses unique analytical challenges as one of the fastest racket sports, characterized by rapid rallies and complex player interactions. Naturally, research has evolved across several key directions, providing a foundation for our current work.

\subsection{Statistical and Computer Vision Approaches}
Computational approaches to badminton analytics have emerged with large-scale data and machine learning. Early works focused on preliminary movement pattern analysis and statistical performance indicators to measure player effectiveness~\cite{gomez2020serving, gomez2020long}. These approaches, while insightful, often relied on descriptive statistics without capturing shot-by-shot dynamics. 

Meanwhile, previous studies developed an end-to-end framework for analyzing broadcast badminton videos through image processing techniques~\cite{ghosh2018towards}, while other research focused on spatiotemporal and stroke feature extraction~\cite{chu2017badminton}. Work in computer vision for badminton has gradually progressed, with efficient systems for shuttlecock tracking \cite{sun2020tracknetv2} and trajectory reconstruction from monocular videos \cite{liu2022monotrack}. These visual tracking systems provide complementary data to the following stroke-level annotations.

\subsection{Data Representation and Datasets}
Recent studies have made significant progress in developing structured representations of badminton data. BLSR (Badminton Language from Shot to Rally) was proposed to formalize badminton data formats from event stream data~\cite{wang2021exploring}. This standardized format enabled more sophisticated analyses by translating match videos into structured datasets. Concurrently, BadmintonDB was released, featuring stroke-level annotations between two players for singles match analysis \cite{ban2022badmintondb}. For performance evaluation and data standardization, comprehensive stroke-level datasets such as ShuttleSet and Shuttleset22 were also introduced~\cite{wang2023shuttleset,wang2024benchmarking}. Shot-by-shot microscopic tactical datasets further supplemented available resources for researchers~\cite{huang2022s}, yet all of these consist only of singles matches.

\subsection{Models for Badminton Game Analysis}
Building upon these foundations, several models have been presented to analyze strategic elements of the game. Deep learning approaches using long short-term dependencies have quantified the influence of shots within rallies~\cite{wang2021exploring}, while position-aware models such as ShuttleNet improved stroke forecasting performance~\cite{wang2022shuttlenet}. These approaches demonstrated the equal importance of temporal dependencies and player-specific characteristics in badminton games. Dynamic graph approaches have also been implemented to improve movement forecasting, with hierarchical fusion techniques~\cite{chang2023where}. RallyNet, a hierarchical offline imitation learning model, tackled the challenge of predicting player behavior in turn-based scenarios using a generative approach~\cite{wang2023generating}. This approach captures players' decision dependencies through contextual Markov decision processes \cite{hallak2015contextual} and employs latent geometric Brownian motion to model player interactions, enabling more realistic imitation of players' decision behaviors. The interpretability of forecasting models was addressed using SHUTTLESHAP~\cite{wang2025shuttleshap} by extending SHAP~\cite{lundberg2017unified} to turn-based game contexts. This approach revealed that player styles significantly influence predictions more than past stroke sequences.

\subsection{Approaches in Other Racket Sports}
While badminton poses unique challenges, similar racket sports like tennis and table tennis have contributed valuable insights applicable to our approach. Tennis analysis has advanced through pose-based stroke recognition techniques~\cite{shah2007automated, ramadan2024detection}, demonstrating the efficacy of pose data for classifying stroke types. Similarly, in table tennis, researchers addressed challenges in player tracking during high-speed exchanges \cite{kulkarni2021table}, utilizing pose keypoints to evaluate player movements. These studies validate the potential of pose-based analysis in racket sports, though they primarily focus on singles play.

The transition from singles to doubles analysis presents significant challenges across racket sports, except for a few attempts in table tennis doubles, which benefit from an alternating stroke rule: players must hit in a predetermined sequence rather than responding dynamically to the ball's location~\cite{liu2025analyzing}. This rule simplifies player tracking and shot attribution in table tennis doubles, as the identity of the next hitter is determined in advance. This simplification does not apply to badminton doubles, where either partner may respond to incoming shuttles based on positioning and strategy, creating significantly more complex player interactions and potential occlusions.

Recent advances in multi-person pose estimation have also contributed to sports analysis capabilities. Niu et al.~\cite{niu2024skeleton} proposed skeleton cluster tracking for robust multi-view multi-person 3D human pose estimation, addressing challenges similar to those encountered in doubles badminton analysis. Additionally, novel approaches using event-based representations for human pose estimation~\cite{yin2024exploring} demonstrate the evolving landscape of pose analysis techniques for fast-paced sports scenarios.

\subsection{Remaining Challenges in Doubles Game}
Despite aforementioned advances, doubles badminton remains rather understudied, with previous research remaining largely unexplored, resulting in a lack of datasets and models for doubles games. Early work on human activity recognition acknowledged the complexity of doubles scenarios~\cite{tran2008human}, yet practical implementation remained focused on singles due to tracking difficulties. Recent work has attempted to address player occlusion issues in doubles through specialized drone-based top-view camera setups~\cite{ding2024estimation}, which provide unobstructed views of all players simultaneously. However, such approaches require specialized equipment and controlled recording environments, limiting their practical applicability to existing broadcast footage or real-time analysis scenarios. Our work aims to bridge this gap by developing a framework that extends singles-trained models to doubles analysis using conventional broadcast camera angles, employing simple but powerful tracking techniques and pose-based classification, making the approach more accessible and broadly applicable.

\begin{figure}[h]
    \centering
    \includegraphics[width=0.6\textwidth]{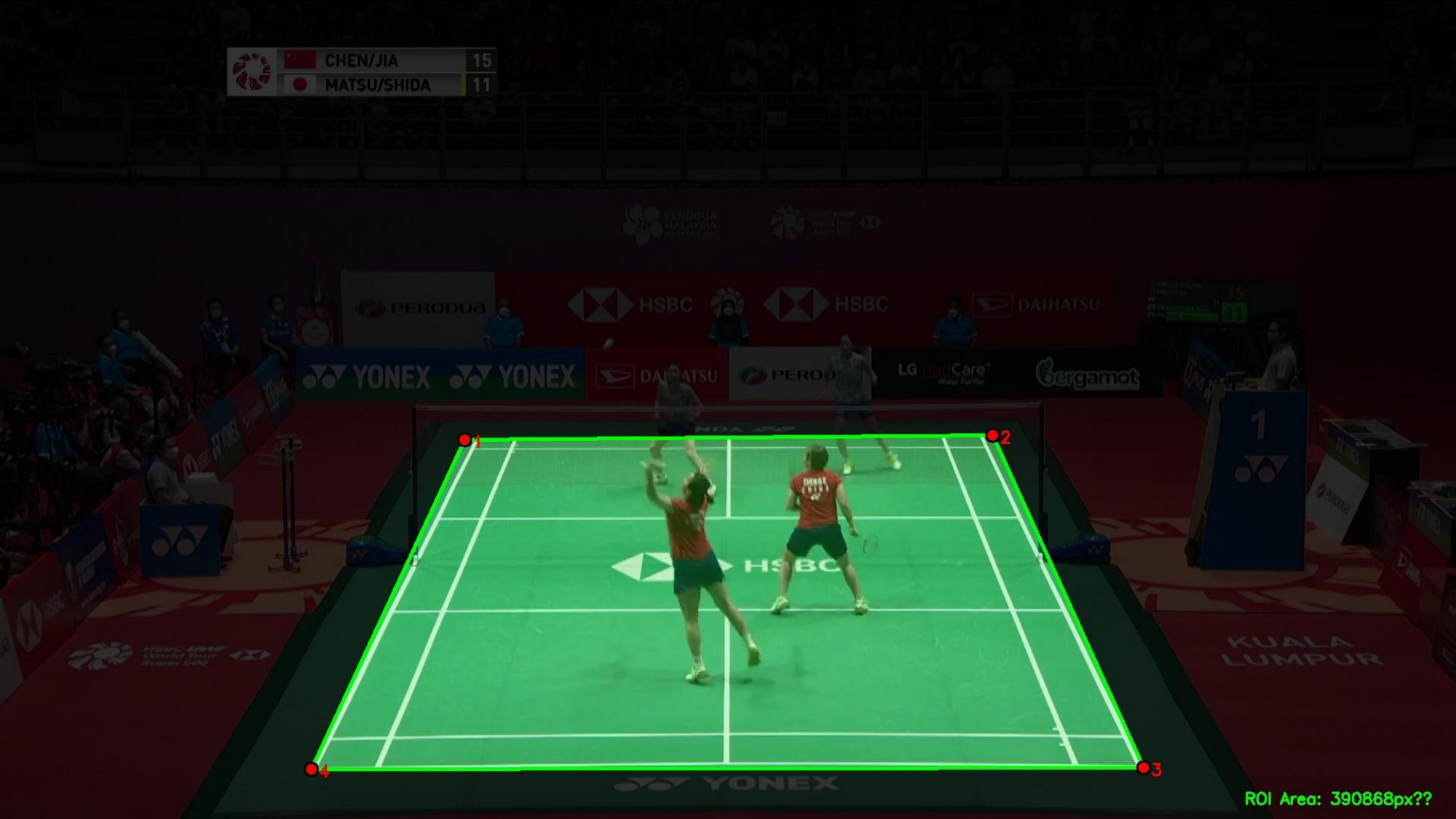}
    \caption{Example of badminton court ROI extraction from overhead camera view.}
    \label{fig:court_roi}
\end{figure}
      
\section{ID Preserving Motion Tracking}\label{sec:tracking}
It is essential to isolate players from other individuals, e.g., referees, umpires, and spectators, to achieve accurate tracking. Therefore, our goal is to identify only the individuals actively playing on the court. First, we employ a pre-trained YOLOv11x model to detect objects classified as ``person''. Since only two or four individuals are players in singles and doubles matches, respectively, we detect the court area from broadcast overhead camera footage. To extract the Region of Interest (ROI) of the badminton court, we utilize a dataset from Roboflow annotated for court corner detection (\url{https://universe.roboflow.com/badminton-rojkf/badminton_court}), which allows us to obtain bounding boxes (bboxes) around the four corners of the court. We train a YOLOv11x model to detect these corner boxes. Using the center point of each bbox as a reference, we select the farthest corner within each bbox and connect these four points to generate the ROI representing the court area, as illustrated in Figure~\ref{fig:court_roi}.

\begin{figure}[h]
    \centering
    \includegraphics[width=\linewidth]{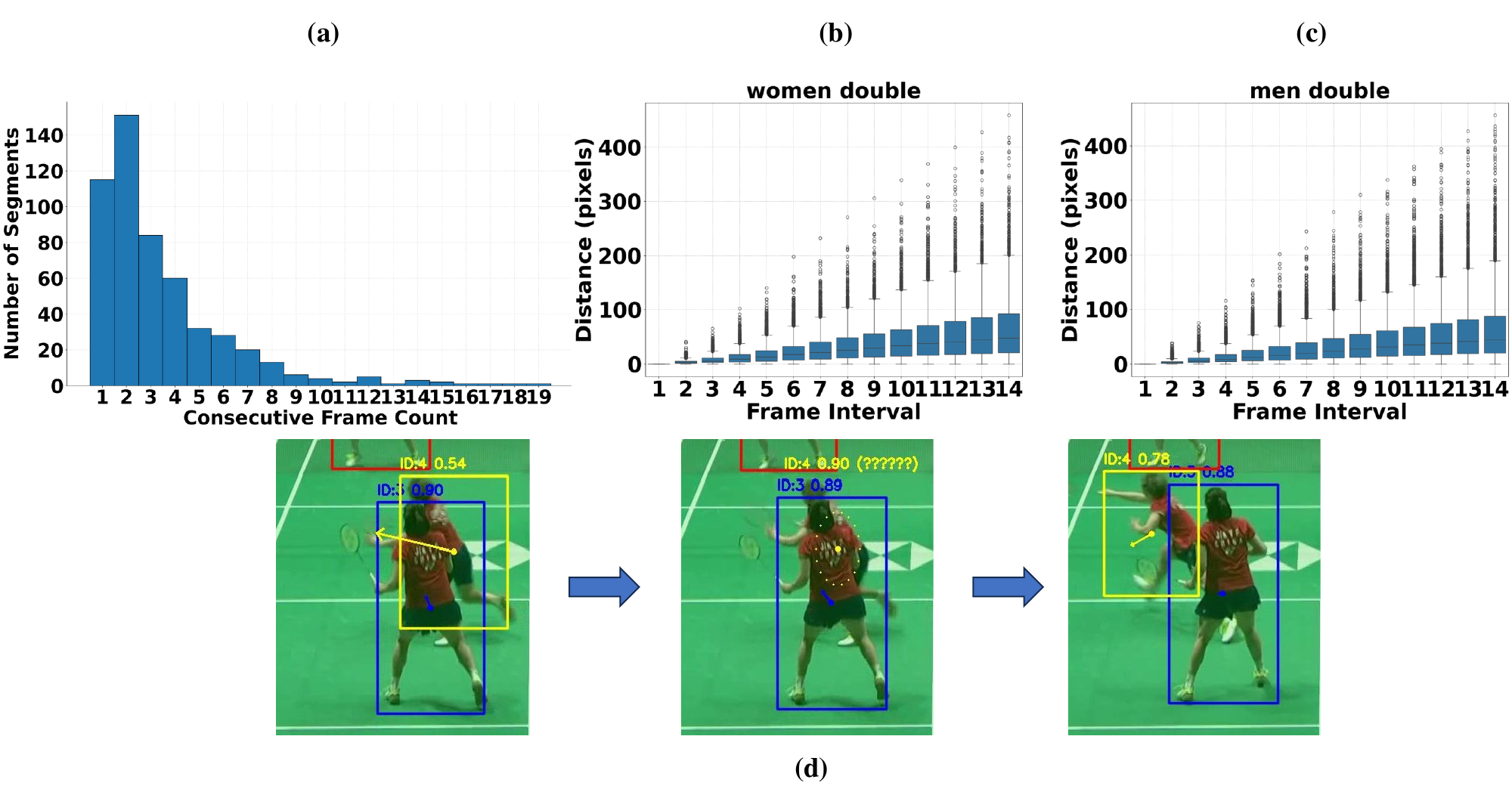}
    \caption{Analysis of predictive tracking performance and ID reassignment. Panel (a) shows the distribution of consecutive missing frames for objects in a sampled men's doubles match. Panels (b) and (c) show the distribution of distances between predicted and actual object centers measured in pixels for men's and women's doubles matches, respectively. For panels (a) and (c), the statistics were calculated from the final round of men's doubles in the Swiss Open 2025 (\url{https://youtu.be/Dnz1JQSgWU4}), whereas panel (b) is derived from the final round of women's doubles in the All England Open 2022 (\url{https://youtu.be/8ucBHScQV-o}). Panel (d) shows a sample visualization of ID reassignment, using the same case as in Figure~\ref{fig:tracking_issue}, illustrating the direction and velocity of an object immediately before disappearance and the reassignment of a new ID after the object reappears.}
    \label{fig:tracking_results}
\end{figure}

To track the identified players, we initially assign IDs using BOT-SORT, the built-in tracking method in YOLOv11x. However, this approach frequently suffers from ID switching issues, especially in badminton doubles matches, as illustrated in Figure~\ref{fig:tracking_issue}. To address this problem, we propose a predictive tracking method based on object velocity. Predictive objects estimate the expected position of a player in the next frame by calculating direction and velocity based on the center coordinates of the object in the previous and current frames: $\vec{S}(t_s + t) = \vec{S}(t_s) + [\vec{S}(t_s) - \vec{S}(t_s-1)] \times t$, where $t_s$ is the time measured in frame numbers of the last successful object detection for a focal player, and $\vec{S}(t)$ is the location of the player at time $t$. Here, $\vec{S}(t_s) - \vec{S}(t_s-1)$ is the velocity of the player at time $t_s$ measured in pixels per frame. When an object reappears within a 200-pixel radius of the predicted object's position, the ID of the predicted object is reassigned to the detected object, effectively mitigating the ID-switching problem (see Figure~\ref{fig:tracking_results}).

We empirically determine a 200-pixel cutoff threshold. First, we find that the missing frame duration is typically less than 10 frames (see Figure~\ref{fig:tracking_results}(a)). Based on this observation, we calculate the Euclidean distances between actual and predicted objects in doubles matches from non-missing consecutive frames. Specifically, we extract 15 consecutive frames without missing objects, then simulate missing frame scenarios by treating consecutive frames from the second frame onwards as missing for durations ranging from 1 to 14 frames, and compare the positions of predicted objects with the actual object positions in these scenarios. We find that the prediction error is typically less than 200 pixels, except for a few outlier cases with long-missing frames (Figure~\ref{fig:tracking_results}(b) and (c)). These results validate our choice of a 200-pixel threshold for ID reassignment, which effectively reassigns the ID for reappearing objects, as illustrated in Figure~\ref{fig:tracking_results}(d).

\begin{table}[h]
\caption{Performance metrics of trained models at optimal accuracy: 1) our Transformer-based model and 2) the YOLOv11x baseline model. When accuracy values are tied, we select the threshold with the higher F1 score.}\label{tab:metric_singles}
\centering
\begin{tabular}{cccccc}
\hline
\textbf{} & \textbf{Thres.} & \textbf{Acc.} & \textbf{Prec.} & \textbf{Recall} & \textbf{F1} \\ \hline\hline
Ours & 0.50 & 0.8617 & 0.8977 & 0.8973 & 0.8975 \\
YOLO & 0.57 & 0.9876 & 0.9880 & 0.9874 & 0.9877 \\
\hline
\end{tabular}
\end{table}

\begin{figure}[h]
    \centering
    \includegraphics[width=0.7\linewidth]{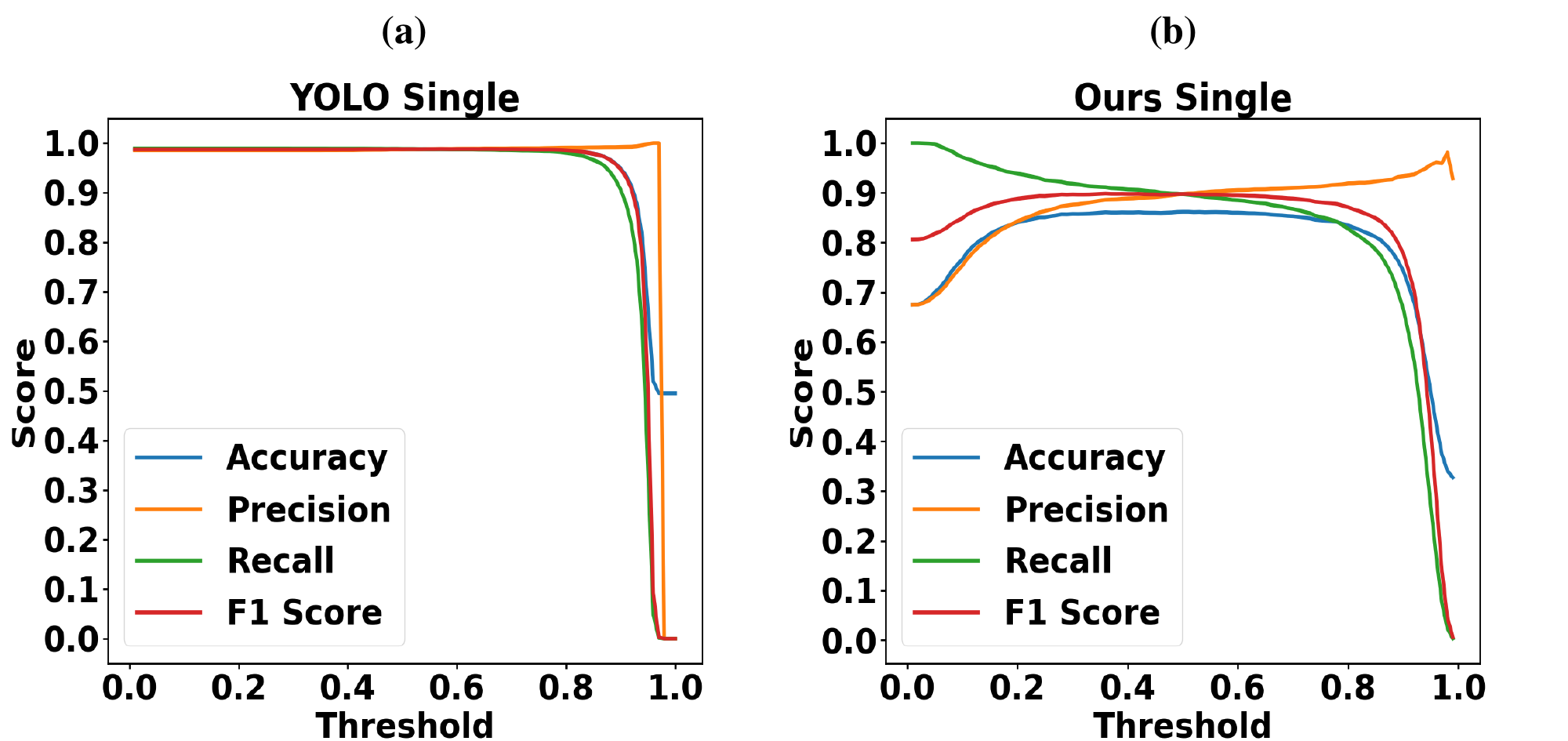}
    \caption{
        Performance metrics of trained models at different thresholds: (a) YOLOv11x baseline model and (b) our Transformer-based model.
    }\label{fig:metric_single}
\end{figure}

\section{ST-GCN Based Backbone Pre-training}\label{sec:gcn}
To learn effective spatiotemporal representations of badminton players' motion, we perform Spatial-Temporal Graph Convolutional Networks (ST-GCN) backbone pre-training based on shot and notshot segments extracted from singles match videos. ST-GCN is well-established for its strong performance in skeleton-based action recognition. First, we collect 40 out of 44 broadcasting videos used in ShuttleSet~\cite{wang2023shuttleset} from the official BWF YouTube channel. The dataset contains documented shot frames that mark the moment when the racket hits the shuttlecock. We excluded four videos for the following reasons: (i) videos were unavailable for download during our study in February 2025 (\textit{Pusarla V. Sindhu Pornpawee Chochuwong HSBC BWF WORLD TOUR FINALS 2020 QuarterFinals} and \textit{NG Ka Long Angus SHI Yu Qi Thailand Masters 2020 SemiFinals}), or (ii) the videos had been updated, causing significant changes in shot frame numbers (\textit{CHOU Tien Chen Jonatan CHRISTIE Sudirman Cup 2019 Quarter finals} and \textit{CHOU Tien Chen NG Ka Long Angus Sudirman Cup 2019 Group Stage}). We define shot segments for each player, where the center frame is annotated as a shot frame in the ShuttleSet dataset. We also define notshot segments, in which the center frame of a video segment is not annotated as a shot frame for each player. Given the turn-based nature of badminton, we select notshot segments from the same temporal window as shot segments, but for the opponent player, because when a certain player hits the shuttlecock, the opponent must not be in the shot state.

\begin{figure*}[h] 
    \centering
    \includegraphics[width=\textwidth]{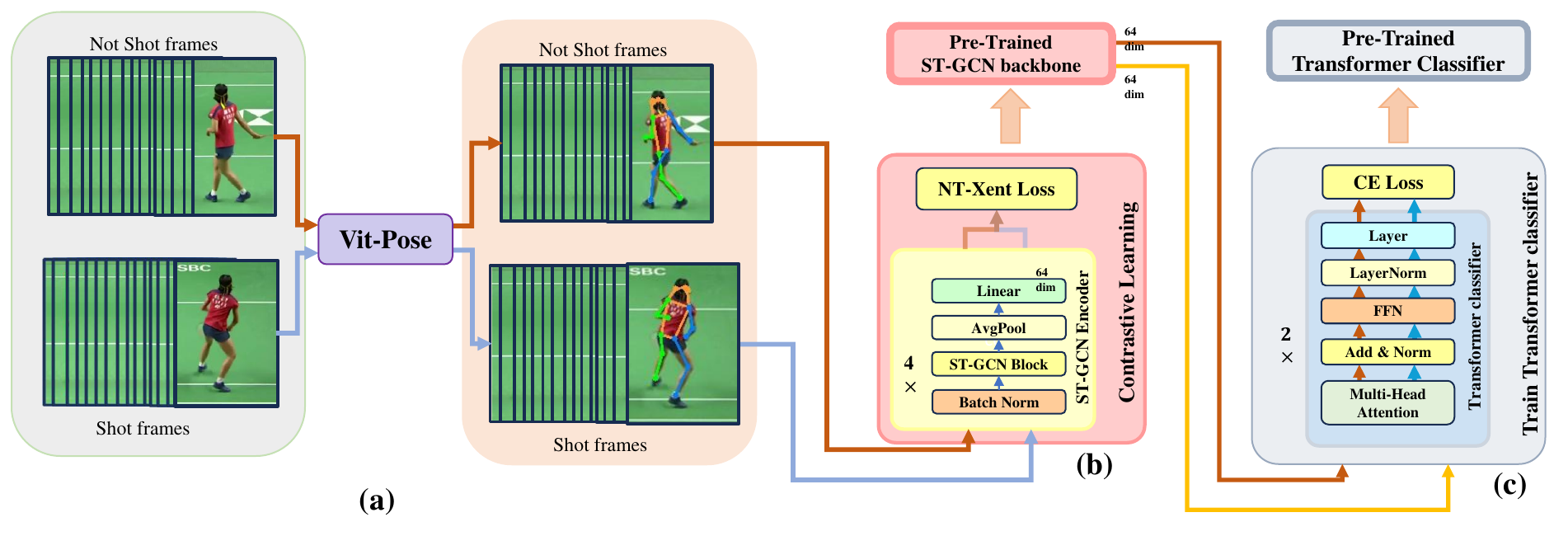}
    \caption{Overview of the training process.
        (a) 17 pose keypoints are extracted from badminton singles match data.
        (b) The extracted keypoints are encoded into a compact 64-dimensional latent space by the ST-GCN encoder, using contrastive learning to learn robust feature representations.
        (c) Using the pre-trained ST-GCN backbone embeddings, a Transformer encoder is trained to classify shot and notshot sequences with cross-entropy loss.}\label{fig:train}
\end{figure*}

We then extract two-dimensional joint coordinates for each frame using ViT-Pose~\cite{xu2022vitpose}. ViT-Pose is a Transformer-based architecture that can effectively capture long-range dependencies compared to simple CNNs. It is suitable for environments such as sports videos where dynamic joint movements frequently occur. We extract 17 two-dimensional joint keypoints in the COCO format (head, shoulders, elbows, wrists, hips, knees, and ankles). We then calculate the relative position of the joints from the midpoint of the two hips, and these relative positions are standardized as $Z=\frac{X-\mu}{\sigma}$ for each coordinate to make them insensitive to resolution changes or physical characteristics such as height and weight. Here, $\mu$ is the mean position of the 17 keypoints, and $\sigma$ is its standard deviation. This relative joint coordinate sequence $\mathbf{S} \in \mathbb{R}^{T \times 34}$ is used as the input to the ST-GCN encoder, where $T$ is the number of frames in the sequence (Figure~\ref{fig:train}(a)).

We employ an ST-GCN encoder consisting of four ST-GCN blocks, each combining graph convolution and temporal convolution in a residual structure (Figure~\ref{fig:train}(b)). This design allows simultaneous modeling of spatial joint connections and temporal motion patterns. The encoder output is reduced to a single vector by average pooling over the frame dimension and then mapped to a 64-dimensional latent embedding $\mathbf{z} \in \mathbb{R}^{64}$ through a linear projection layer. The 64-dimensional size is empirically selected based on the trade-off between model complexity and expressive power, following previous studies. To train the embedding space, we use contrastive loss NT-Xent~\cite{chen2020simple}. Since players on opposite sides of the net face different directions, we train two separate backbone models for front-court and back-court players. To address the imbalance between left-handed and right-handed players, we also train on horizontally flipped versions of all videos. During backbone training, we use all segments extracted from the 40 videos for training. We apply early stopping when the loss does not decrease for 15 consecutive epochs, and model parameters at the point of the lowest recorded loss are saved as a checkpoint for the backbone.

\section{Transformer Based Shot Frames classification}\label{sec:transformer}
We then train shot classifiers for singles using a 7:3 train-test split (Figure~\ref{fig:train}(c)). The 64-dimensional embeddings obtained from the pre-trained ST-GCN backbone are used as inputs to a Transformer encoder for binary classification between shot and notshot segments. For baseline comparison, we also train a vision-only model using YOLOv11x directly on shot frames without skeletal keypoint extraction. The Transformer encoder effectively captures temporal dependencies through its self-attention mechanism, enabling the model to focus on relevant motion cues distributed across the sequence without being limited by local receptive fields. Since players on opposite sides of the net exhibit different motion patterns, we train separate models for front-court and back-court players relative to the camera perspective for both our model and the YOLO baseline. This approach allows the classifiers to discern subtle temporal patterns specific to each court position. We optimize both classifiers using cross-entropy loss, which is well-suited for binary classification. Integrating contrastively learned embeddings and Transformer-based temporal modeling leads to robust and efficient shot classification performance.

Table~\ref{tab:metric_singles} and Figure~\ref{fig:metric_single} show the performance comparison between our model and the YOLOv11x baseline on singles badminton data. At their respective optimal thresholds, the YOLOv11x baseline achieves higher performance with an accuracy of 0.9876 (threshold 0.57), compared to our model's accuracy of 0.8617 (threshold 0.50). The baseline model also achieves consistently higher scores across all other metrics. The threshold-performance curves (Figure~\ref{fig:metric_single}) reveal distinct patterns between the two approaches. The YOLOv11x baseline (Figure~\ref{fig:metric_single}(a)) maintains stable high performance across a wide range of thresholds from 0.01 to 0.70, with all metrics remaining above 0.90, before experiencing a sharp decline beyond threshold 0.80. In contrast, our Transformer model (Figure~\ref{fig:metric_single}(b)) shows more gradual performance variations across thresholds, with optimal performance occurring at lower threshold values around 0.5. While the vision-only YOLOv11x baseline achieves higher performance on singles data, our approach offers the advantage of transferability to doubles scenarios without requiring additional training, as we demonstrate in the next section.

\section{Applying Singles-trained Model to Doubles}
\begin{figure*}[h]
    \centering
    \includegraphics[width=\textwidth]{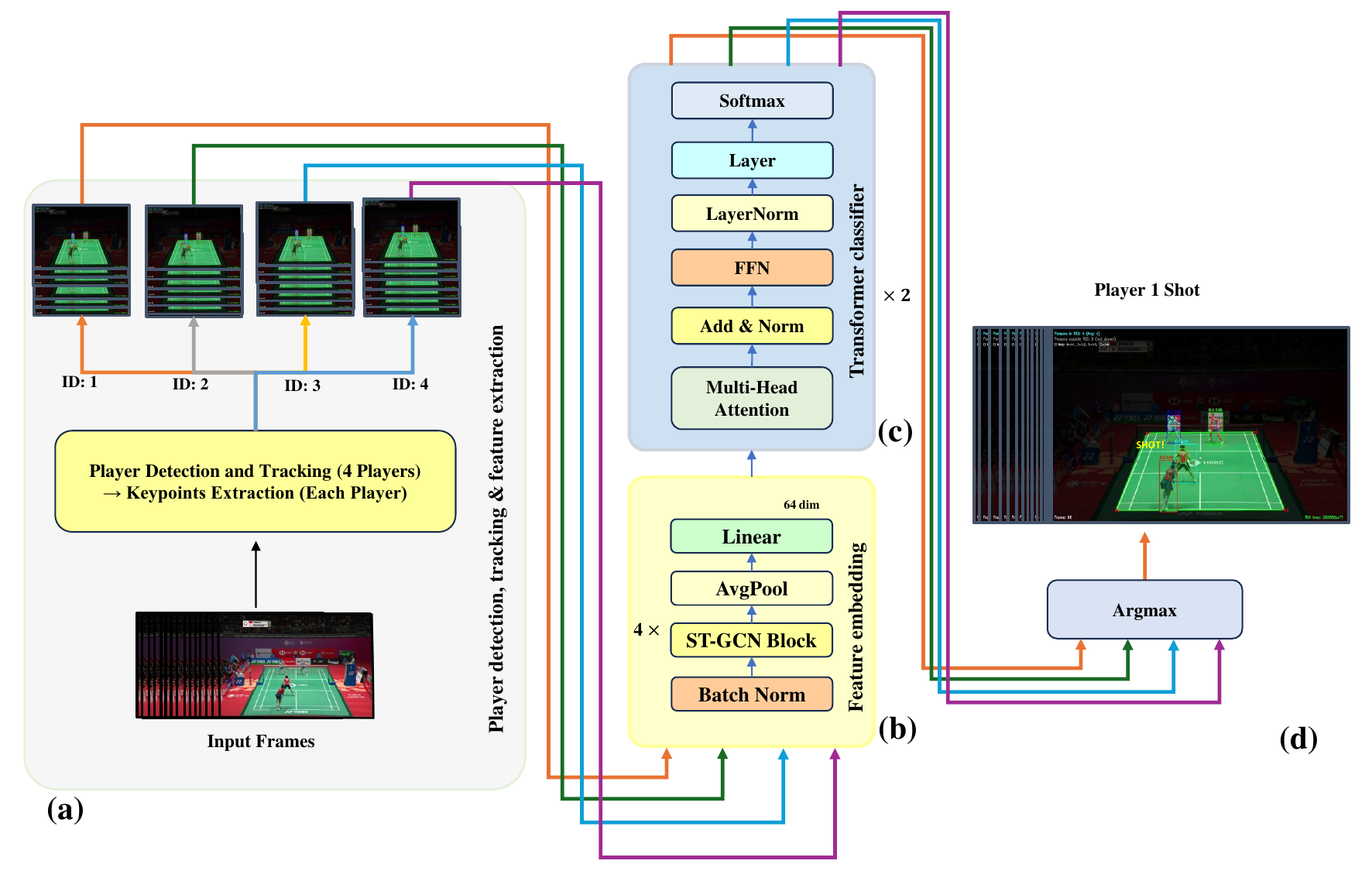}
    \caption{
        Schematic overview of the inference process. 
        (a) A 15-frame clip is input to an object detection and tracking module to localize four players, assign unique IDs, and extract their skeletal keypoints.
        (b) The keypoints are encoded by the ST-GCN backbone, pre-trained with contrastive learning, and projected into a 64-dimensional feature space through a linear layer.
        (c) A two-layer Transformer encoder classifier predicts the binary label ``Shot'' or ``Not Shot'' for each player. The models in (b) and (c) are trained on singles videos (see Sections~\ref{sec:gcn}~and~\ref{sec:transformer}).
        (d) The model selects the player with the highest confidence score among the four predictions as the shot player. If all players fall below the tunable confidence threshold, the model concludes that no shot has been made.
    }\label{fig:inference_process}
\end{figure*}

To test the generalization capability of our singles-trained model, we apply it directly to doubles badminton matches without additional training. The inference pipeline remains identical to the singles application, as illustrated in Figure~\ref{fig:inference_process}. A 15-frame clip is processed through the same object detection and tracking module to localize four players in doubles matches, assign unique IDs, and extract skeletal keypoints for each player (Figure~\ref{fig:inference_process}(a)). The key adaptation lies in scaling the system from tracking two players in singles to four players in doubles. Our predictive tracking method, originally designed to handle occlusion between two players, effectively manages the increased complexity of four-player interactions. The skeletal keypoints extracted from each player are independently processed through the pre-trained ST-GCN backbone, which projects them into the same 64-dimensional latent space learned from singles data (Figure~\ref{fig:inference_process}(b)).

Although trained only on singles matches, the ST-GCN backbone demonstrates transferability to doubles scenarios. The learned spatiotemporal representations successfully capture fundamental badminton motion patterns that are consistent across different game formats. Each player's keypoint sequence is encoded independently, enabling the model to handle multiple players simultaneously without architectural modifications. The Transformer encoder classifier processes each player's embedding separately, predicting binary shot/notshot labels for all four players (Figure~\ref{fig:inference_process}(c)). The final decision employs the same confidence-based selection mechanism: the player with the highest confidence score among the four predictions is identified as the shot player, or no shot is detected if all players fall below the threshold (Figure~\ref{fig:inference_process}(d)). If a shot player is identified, the frame is classified as a shot frame; otherwise, it is classified as a notshot frame, thereby enabling effective binary frame-level classification.

\begin{figure*}[h]
    \centering
    \includegraphics[width=\textwidth]{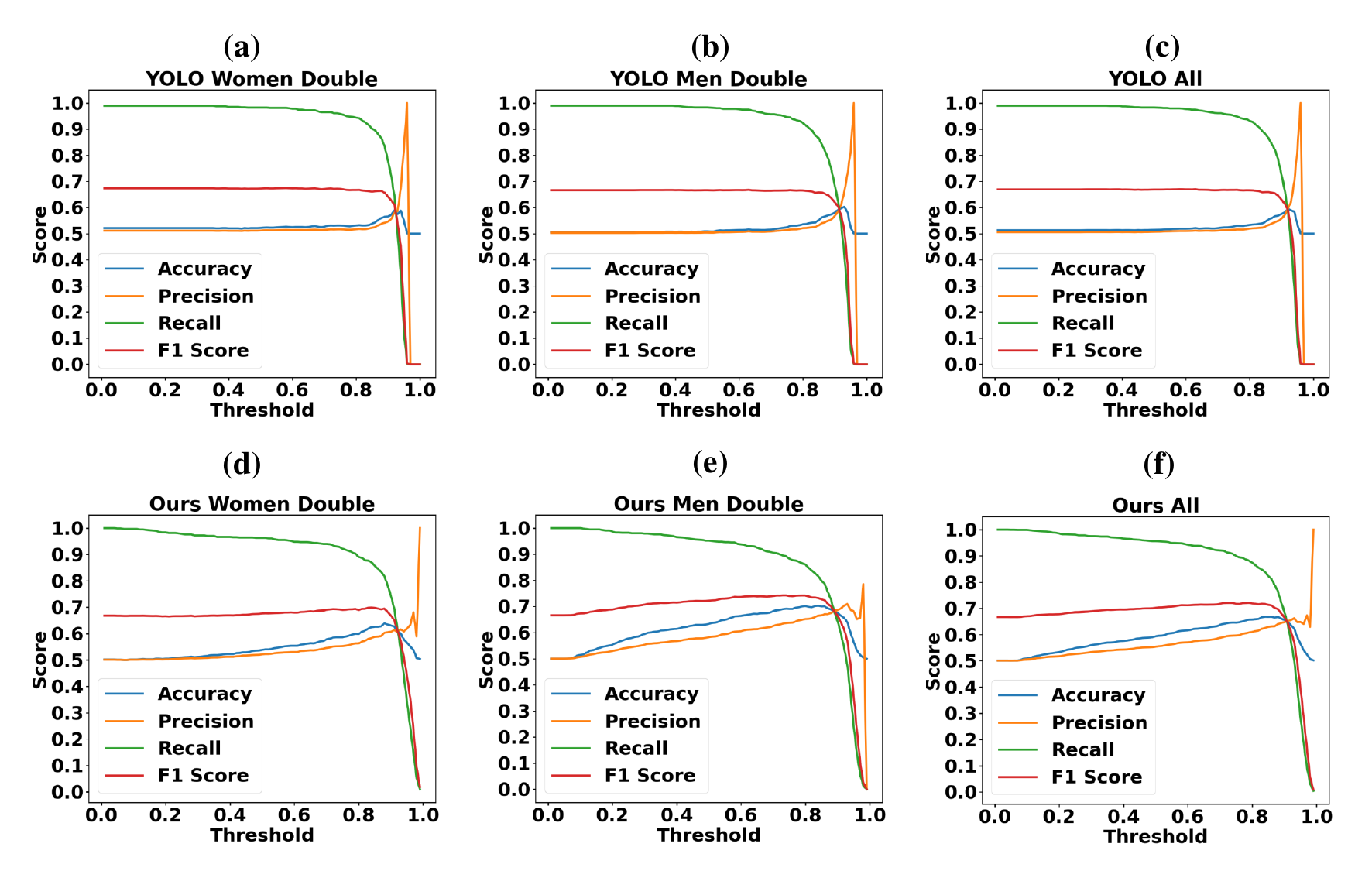}
    \caption{
        Performance metrics of trained models at different thresholds for the doubles validation set: (a)--(c) YOLOv11x baseline model and (d)--(f) our Transformer-based model. For validation, we manually annotated one women's doubles match and one men's doubles match. The ``all'' category represents the combined dataset of both doubles matches.}\label{fig:metric_double}
\end{figure*}

The validation results reveal a significant performance reversal compared to the singles evaluation (Table~\ref{tab:metric_double} and Figure~\ref{fig:metric_double}). The validation datasets are from the same matches used in Figure~\ref{fig:tracking_results} for tracking analysis, yielding 1,698 and 1,276 shot/notshot frames for the men's and women's games, respectively. By design, this validation dataset contains an equal number of shot and notshot frames. Interestingly, our approach demonstrates twice the domain transfer robustness (19.3\% vs 39.9\% degradation), demonstrating that skeletal representations are inherently more transferable across game formats than raw visual features. Our Transformer-based model achieves an overall accuracy of 0.6686 on the combined doubles dataset (threshold 0.86), while the YOLOv11x baseline drops dramatically to 0.5941 (threshold 0.92).

Gender-specific analysis reveals distinct transferability patterns. Our model maintains consistent performance across genders, with men's doubles achieving 0.7020 accuracy (threshold 0.84) and women's doubles reaching 0.6389 (threshold 0.88). In contrast, the YOLOv11x baseline shows severe degradation, particularly for women's doubles (0.5893) compared to men's doubles (0.6019), with sharp recall drops near the accuracy threshold, indicating poor generalization to multi-player environments. Our model maintains high recall ($81.24\%$), successfully detecting over $80\%$ of actual shots across a wide range of thresholds (0.1-0.9; Figure~\ref{fig:metric_double}). This is valuable for sports analytics, where missing critical moments (false negatives) is more problematic than false detections (false positives), such as in broadcast highlights or training analyses where capturing all shot events is paramount.

\begin{table}[h]
\caption{Performance metrics of trained models at optimal accuracy for the doubles validation set: (a)--(c) our Transformer-based model and (d)--(f) YOLOv11x baseline model. For validation, we manually annotated one women's doubles match and one men's doubles match. The ``all'' category represents the combined dataset of both doubles matches.}\label{tab:metric_double}
\centering
\begin{tabular}{cccccc}
\hline
\textbf{} & \textbf{Thres.} & \textbf{Acc.} & \textbf{Prec.} & \textbf{Recall} & \textbf{F1} \\ \hline\hline
Ours (ALL) & 0.86 & 0.6686 & 0.6311 & 0.8124 & 0.7104 \\
Ours (Men) & 0.84 & 0.7020 & 0.6644 & 0.8163 & 0.7326 \\
Ours (Women) & 0.88 & 0.6389 & 0.6028 & 0.8182 & 0.6941 \\ \hline
YOLO (ALL) & 0.92 & 0.5941 & 0.5962 & 0.5837 & 0.5899 \\
YOLO (Men) & 0.93 & 0.6019 & 0.6599 & 0.4205 & 0.5137 \\
YOLO (Women) & 0.92 & 0.5893 & 0.5805 & 0.6442 & 0.6107 \\
\hline
\end{tabular}
\end{table}

These results demonstrate that despite the superior performance of the baseline vision model on singles data, our approach demonstrates significantly better cross-format generalization. The skeletal keypoint-based representation with contrastive pre-training proves more robust to the increased complexity of doubles play, while the vision-only baseline struggles to adapt to multi-player environments without additional training. This cross-format transferability demonstrates the potential of transfer learning in sports action recognition, where collecting labeled data for every scenario can be prohibitively expensive.

\section{Discussion}
In this study, we demonstrate that singles-trained ST-GCN models can effectively transfer to doubles badminton analysis without requiring extensive doubles training data. This cost-efficient approach achieves functional performance while significantly reducing annotation costs, making it practical for sports analytics applications where comprehensive labeling is expensive. The successful cross-format transfer demonstrates that fundamental spatiotemporal motion patterns in badminton exist beyond specific game configurations.

While applying singles models to doubles scenarios introduces additional complexity, our pose-based approach exhibits remarkable cross-format resilience. The accuracy transition from 0.8617 (singles) to 0.6686 (doubles) demonstrates superior transferability compared to vision-only methods, which suffer twice the degradation. This resilience reflects the fundamental advantage of skeletal representations in capturing motion patterns that transcend specific game configurations. Gender-based performance disparities also emerge, with men's doubles achieving 0.7020 accuracy at threshold 0.84 versus women's doubles reaching 0.6389 at threshold 0.88. This gap may stem from training data imbalance (30 male vs. 10 female videos) despite normalization efforts, or from fundamental biomechanical differences between genders that we did not account for: men's doubles typically feature more power-based shots with pronounced motion signatures, while women's doubles emphasize finesse and placement strategies that produce subtler movement patterns. Additional limitations include dependency on broadcast camera angles and threshold optimization challenges across different player demographics. The higher optimal threshold in doubles (0.86 vs. 0.55) indicates that multi-player environments create more ambiguous patterns, requiring more conservative detection strategies.
Despite these limitations, our work has significant implications for sports research. We provide the first systematic evaluation of cross-format transfer learning in badminton action recognition, establishing a practical baseline for cost-effective model deployment. Our predictive tracking method effectively addresses the critical ID switching problem in multi-player scenarios, with the empirically determined $200$-pixel threshold proving robust for typical badminton court dynamics. This lightweight post-processing approach maintains computational efficiency while preserving identity consistency, which is essential for temporally coherent pose sequence analysis. Our findings regarding the limited predictability of vision-only approaches indicate that incorporating biomechanical theories is required to build more generalizable models for sports analysis. The skeletal keypoint-based representation captures motion abstractions that remain consistent across game formats, unlike vision-only methods that struggle with increased visual complexity and player occlusions in multi-player environments. The demonstrated transferability opens possibilities for rapid deployment across various sports beyond badminton without extensive retraining.

Our approach provides a practical framework for developing cost-effective sports analytics solutions with broad applicability, demonstrating that pose-based models can bridge domain gaps that vision-only approaches cannot. This work may inspire further research in cross-domain transfer learning for sports analysis and contribute to making advanced machine learning tools more accessible to the sports research community.

The demonstrated transferability of our pre-trained ST-GCN backbone models, despite being trained only on singles match data, suggests that the learned spatiotemporal representations capture fundamental badminton motion patterns applicable across different game formats. This opens several promising applications for future work. The backbone can be extended to recognize different shot types (e.g., smash, drop, clear, drive) by training additional classification heads, enabling fine-grained shot analysis. When combined with homography transformation, our tracking system can also determine shot positions on the court, providing spatial context for tactical analysis. The temporal modeling capabilities can support advanced tactical analysis, such as identifying rally patterns or analyzing player positioning strategies. Additionally, the approach can be adapted for real-time match analysis and coaching assistance systems. We anticipate that the public availability of these resources will accelerate research in sports analytics and benefit the broader computer vision community working on complex human action recognition.

\section*{CRediT authorship contribution statement}
\textbf{Seungheon Baek}: Conceptualization, Methodology, Software, Validation, Formal analysis, Investigation, Data curation, Writing – original draft, Visualization. \textbf{Jinhyuk Yun}: Conceptualization, Methodology, Formal analysis, Resources, Writing – review \& editing, Visualization, Supervision, Project administration, Funding acquisition.

\section*{Declaration of Competing Interest}
The authors declare that they have no known competing financial interests or personal relationships that could have appeared to influence the work reported in this paper.

\section*{Data Availability}
All training code, model architectures, and pre-trained weights for both the ST-GCN backbone and the Transformer-based shot classifier are available at GitHub (\url{https://github.com/100-heon/badminton_double_analysis}). A demonstration video of the current system is available at YouTube (\url{https://youtu.be/sqHbnD4Ig2g}). The ShuttleSet~\cite{wang2023shuttleset} dataset, which contains documented shot frames that mark the moment when the racket hits the shuttlecock, is available from its official GitHub project page (\url{https://github.com/wywyWang/CoachAI-Projects/tree/main/ShuttleSet}), with corresponding videos downloadable from the official BWF YouTube channel.

\section*{Acknowledgments}
This research was supported by the MSIT (Ministry of Science and ICT), Republic of Korea, under the Innovative Human Resource Development for Local Intellectualization support program (IITP-2025-RS-2022-00156360, 30\%) supervised by the IITP (Institute for Information \& Communications Technology Planning \& Evaluation). This work was also supported by the National Research Foundation of Korea (NRF) funded by the Korean government 2022R1A2C1091324). This research was also supported by the Global Humanities and Social Sciences Convergence Research Program through the National Research Foundation of Korea(NRF), funded by the Ministry of Education (2024S1A5C3A02042671). The Korea Institute of Science and Technology Information (KISTI) also supported this research by providing KREONET, a high-speed Internet connection.

\section*{Declaration of generative AI and AI-assisted technologies in the writing process}
During the preparation of this work the authors used Grammarly (\url{https://www.grammarly.com/}), Gemini (\url{https://gemini.google.com/}), and Claude (\url{https://claude.ai/}) in order to correct grammatical errors and refine the phrasing in this paper. The authors also used Claude (\url{https://claude.ai/}) and ChatGPT (\url{https://chatgpt.com/}) in order to assist with writing some of the source code. After using these tools/services, the authors reviewed and edited the content as needed and take full responsibility for the content of the published article.



\bibliography{references}

\end{document}